\documentclass[conference]{IEEEtran}
\usepackage{cite}
\usepackage{physics, amsmath,amssymb,amsfonts}
\usepackage{graphicx}
\usepackage{textcomp}
\usepackage{xcolor}
\usepackage{hyperref}
\usepackage{booktabs}
\usepackage{caption}
\usepackage{subcaption}
\usepackage{footmisc}
\usepackage{fancyhdr}

\def\BibTeX{{\rm B\kern-.05em{\sc i\kern-.025em b}\kern-.08em
    T\kern-.1667em\lower.7ex\hbox{E}\kern-.125emX}}

\newcommand\extrafootertext[1]{%
    \bgroup
    \renewcommand\thefootnote{\fnsymbol{footnote}}%
    \renewcommand\thempfootnote{\fnsymbol{mpfootnote}}%
    \footnotetext[0]{#1}%
    \egroup
}

\usepackage{algorithm}
\usepackage{algpseudocode} 
\graphicspath{ {img/} }

\newcommand{\biocop}{BIOCOP2008 }
\newcommand{\trimodal}{MIT LL TRIMODAL}
\definecolor{darkgreen}{rgb}{0.0, 0.5, 0.0}
\begin{document}

\IEEEoverridecommandlockouts

\fancypagestyle{plain}{
    \fancyhf{}
    \fancyhead[L]{\textcolor{red}{M. R. Dale, E. Singer, B. J. Borgström, A. Ross, “To Impute or Not: Recommendations for Multibiometric Fusion,” Proc. of IEEE International Workshop on Information Forensics and Security (WIFS), (Nuremberg, Germany), December 2023.}}
    \renewcommand{\headrulewidth}{0pt} 
}

\pagestyle{plain}

\title{To Impute or Not: Recommendations for Multibiometric Fusion}






\author{
    \IEEEauthorblockN{Melissa R Dale\IEEEauthorrefmark{1}, Elliot Singer\IEEEauthorrefmark{2}, Bengt J. Borgstr\"{o}m\IEEEauthorrefmark{2}, Arun Ross\IEEEauthorrefmark{1}}
    \IEEEauthorblockA{\IEEEauthorrefmark{1}\textit{Dept of Computer Science and Engineering} \\
    \textit{Michigan State University}\\
    East Lansing, MI, USA 
    \\}
    \IEEEauthorblockA{\IEEEauthorrefmark{2}\textit{Artificial Intelligence Technology and Systems Group} \\
    \textit{MIT Lincoln Laboratory}\\
    Lexington, MA, USA 
    \\}
}

\maketitle

\thispagestyle{plain}
\begin{abstract}


Combining match scores from different biometric systems via fusion is a well-established approach to improving recognition accuracy. However, missing scores can degrade performance as well as limit the possible fusion techniques that can be applied.  Imputation is a promising technique in multibiometric systems for replacing missing data. In this paper, we evaluate various score imputation approaches on three multimodal biometric score datasets, viz. NIST BSSR1, BIOCOP2008, and \trimodal, and investigate the factors which might influence the effectiveness of imputation. Our studies reveal three key observations: (1) Imputation is preferable over not imputing missing scores, even when the fusion rule does not require complete score data. (2) Balancing the classes in the training data is crucial to mitigate negative biases in the imputation technique towards the under-represented class, even if it involves dropping a substantial number of score vectors. (3) Multivariate imputation approaches seem to be beneficial when scores between modalities are correlated, while univariate approaches seem to benefit scenarios where scores between modalities are less correlated.

\end{abstract}


\section{Introduction}
\label{sec:Intro}

\extrafootertext{DISTRIBUTION STATEMENT A. Approved for public release. Distribution is unlimited.  This material is based upon work supported under Air Force Contract No. FA8702-15-D-0001. Any opinions, findings, conclusions or recommendations expressed in this material are those of the author(s) and do not necessarily reflect the views of the U.S. Air Force.  Delivered to the U.S. Government with Unlimited Rights, as defined in DFARS Part 252.227-7013 or 7014 (Feb 2014). Notwithstanding any copyright notice, U.S. Government rights in this work are defined by DFARS 252.227-7013 or DFARS 252.227-7014 as detailed above. Use of this work other than as specifically authorized by the U.S. Government may violate any copyrights that exist in this work.}

Biometric systems are indispensable for recognizing individuals based on the uniqueness of their biological and behavioral traits such as face, fingerprint, iris, voice, and gait \cite{ross2006handbook}. However, in many real-world applications, the reliance on a single biometric modality may not be sufficient to meet the criteria of high recognition accuracy and enhanced security. As a result, the fusion of multiple biometric modalities or algorithms has become a crucial avenue of research and development. In addition to improving performance and increasing security \cite{brunelli1995person, prabhakar2002decision, toh2004exploiting, ross2003information}, using multiple biometrics can also improve accessibility.  By incorporating different biometric modalities, biometric systems can accommodate individuals with varying physical characteristics or limitations. For example, individuals who may have difficulty providing a fingerprint due to a physical disability can still participate in the system by utilizing their face or voice as an alternative modality \cite{boulgouris2009biometrics, mishra2010multimodal}. However if the fusion approach is not designed carefully, the benefits of fusion can be stymied by an unnecessarily convoluted system with slow performance \cite{allano2010tuning}.

Biometric fusion can be accomplished at multiple levels, including data, feature, score, decision, and rank levels. In score-level fusion, the match scores from the participating modalities or matchers are combined.  One design decision when considering score-level fusion is how to handle missing match score values. Missing scores can arise from various factors within a biometric system. This includes failures in acquiring the biometric sample or encountering samples of insufficient quality. Additionally, the integration of new biometric modalities into an existing system may introduce a discrepancy where the input probe data contains more modalities than the corresponding gallery identities, resulting in a missing data scenario. While some fusion techniques can be applied to data with missing values, such as the simple sum rule, \footnote{A score vector consists of scores from multiple matchers.} many fusion techniques, however, are not designed to implicitly account for missing scores. In these instances, a choice must be made: either ignore the score vectors that contain missing values or replace missing values with an estimated value (a process known as imputation). If the proportion of missing data is small, simply ignoring those samples may not influence overall aggregate performance. However, if there is a large proportion of missing data, ignoring incomplete data may be harmful to the performance. Imputation can help address these situations. Additionally, it has been shown that implementing imputation techniques can improve biometric recognition performance even when not required by the fusion rule \cite{DaleMissing}. The authors in \cite{DaleMissing} show that applying the simple sum rule with the imputed score data frequently improved both verification and identification performance in biometric recognition tasks even when up to 90\% of data was incomplete. However, imputation can also add undesirable computational and time complexity to a multibiometric system.


\section{Background}
\label{sec:Background}

Fusion in multibiometrics encompasses various levels of integration, including data, feature, score, decision, or rank \cite{ross2006handbook}. Among these, score level fusion has received significant attention due to its applicability when working with biometric systems that provide match scores rather than raw data or features. 


The simple sum rule is a popular choice of fusion thanks to its straight forward approach that often produces strong recognition accuracies. Simple sum fusion is a transformation-based approach, i.e, since all scores must share a common range, a transformation is required (e.g., normalizing scores into the range of [0-1]).  Other approaches to score-level fusion include classifier-based techniques \cite{fierrez2005discriminative, ma2005classification} and density-based techniques \cite{nandakumar2007likelihood}. These techniques often require the estimation of a number of parameters and, hence, depend on the availability and representativeness of the training data. 


When a fusion technique requires score data to be complete, careful consideration must be paid to \textbf{\textit{how}} the data is missing. For this analysis, we define a score vector as the vector of match scores between identities $i$ and $j$, where $\vectorbold{s_{ij}} = [s_1, ..., s_m]$ for the $m$ modalities present.

Rubin defines the following patterns of missing values \cite{rubin1976inference}: Missing Completely at Random (MCAR), Missing at Random (MAR), and Missing Not at Random (MNAR). Accurately identifying the reason for missing data is vital, as the suitability of imputation methods depends on whether the missingness follows the MCAR or MAR assumptions. It is important to note that MNAR can introduce biases and lead to erroneous conclusions if not properly addressed. Assuming the missingness is either MCAR or MAR, the literature provides various approaches for handling missing data, which are outlined in the following paragraphs.

One option for missing data is to simply ignore vectors with missing scores. This approach is referred to as \textit{Listwise Deletion} \cite{kang2013prevention} and works if there is only a small proportion of missing data and that missing data is truly MCAR. If data is missing because of a failing sensor, for instance, the missing values are due to the sensor and, thus, a bias would be introduced into the analysis. A drawback of this approach is that otherwise usable score vectors are lost.

Imputation is an alternative approach to listwise deletion. A univariate approach to imputation is to simply replace missing values with the corresponding modality's mean or median score observed in the training data. This approach is referred to as univariate because imputation is only dependent on one modality and is unaffected by other modalities. For example, consider a scenario presented in Table \ref{tab:demo}. Median imputation, for example, would replace the face modality's missing score with $0.41$ (the median of the available face modality scores from the face scores in the training data) and the missing fingerprint score would be replaced with $0.74$. 


\begin{table}[htbp]
\centering
\caption{A simple example of a score dataset with missing values, denoted as ?.}
\begin{tabular}{|c|c|c|c|}
\hline 
 \textbf{Subject} & \textbf{Face} &  \textbf{Fingerprint} &  \textbf{Iris} \\\hline 

Subject 1 &  ? &         0.74 &  1.00 \\\hline 
Subject 2 & 0.41 &         0.89 &  0.47 \\\hline 
Subject 3 & 0.27 &          ? &  0.03 \\\hline 
Subject 4 & 0.85 &         0.00 &  0.31 \\\hline 

\end{tabular}
\label{tab:demo}
\end{table}

Another imputation approach leverages potential relationships between modality scores to better estimate the missing value. One such multivariate approach is \textit{Multivariate Imputation by Chained Equations} (MICE), where missing values are temporarily filled with a placeholder value and then iteratively updated using a trained machine learning model \cite{van2011mice}. 

In the given example, shown in Table \ref{tab:demo}, both the face and iris missing values are initialized with each modality's mean or median. The scores of individual modalities are sequentially and iteratively updated with a specified machine learning classifier. Once the classifier has been trained, the missing values are updated from the initial placeholder value to the value predicted by the trained classifier, and then the next modality's scores are fixed and the classifier is trained again to update the placeholder values. This process is repeated for a specified number of iterations, or until the imputed values stop changing between iterations. 

Previous studies have shown that using imputation methods can reliably improve recognition performance in multibiometric systems when scores are missing \cite{ding2012comparison, DaleMissing}. Alternatively, fusion approaches that can adapt to missing data points, such as a likelihood ratio scheme that incorporates both rank and score, can also be used \cite{nandakumar2009fusion, fatukasi2008estimation}.



\section{Approach}
The experiments in this paper are conducted on 3 multimodal biometric datasets (described in detail below). It should be noted that only match scores are available in these datasets, and no additional information about the samples or the sample quality is known. 

For the \trimodal{} dataset, scores are distributed in development (dev) and test sets.  In this dataset, the training phase is conducted on the dev set, which comprises less than 10\% of the total \trimodal {} dataset. The remaining two datasets are randomly divided into train (80\%) and test (20\%) sets. All dataset partitions are subject disjoint, ensuring that the data used for training and testing do not overlap, and all imputation approaches applied to the test set are exclusively derived from calculations performed on the training set.

We next simulate up to 90\% incomplete score vectors on 3 versions of each dataset: randomly missing across all score vectors, randomly missing from genuine score vectors, and randomly missing from  imposter score vectors. To ensure the robustness of our findings, we repeat this simulation process five times on the complete training and testing partitions of the datasets.   

We finally compare the performance of applying the simple sum fusion on the non-imputed version to the imputed versions, noting the mean Pearson's pairwise and Spearman's Rank correlation coefficients between scores of all possible modality pairs in the dataset. This experimental setup is summarized in Table \ref{tab:experiments}.

\begin{table}[htbp]
\centering
\caption{Summary of settings used in the experiments.}
\resizebox{\columnwidth}{!}{%
\begin{tabular}{|l|l|}
\hline 
\textbf{Experimental Parameter} & \textbf{Settings} \\ \hline
\textbf{Multibiometric Datasets}  &  \begin{tabular}[c]{@{}l@{}}NIST BSSR1 \cite{BSSR1nist}\\ \biocop \\ \trimodal \cite{mitlltrimodal} \end{tabular}\\ \hline
\textbf{Training, Testing Split} & \begin{tabular}[c]{@{}l@{}}80\%, 20\% (NIST BSSR1, \biocop)\\ 7\%, 93\% (\trimodal) \end{tabular} \\ \hline
\textbf{\# of Missing Score Simulations} & 5 \\ \hline
\textbf{\% Missing} & \begin{tabular}[c]{@{}l@{}}{[}0, 10, 20, 30, 40,\\ 50, 60, 70, 80, 90{]}\end{tabular} \\ \hline
\textbf{Univariate Imputations} & \begin{tabular}[c]{@{}l@{}}Mean\\ Median\end{tabular} \\ \hline
\textbf{Multivariate Imputations} & \begin{tabular}[c]{@{}l@{}}Bayesian Regression \cite{mackay1992bayesian}\\ Decision Tree \cite{breiman2017classification} \\ K Nearest Neighbors \cite{kramer2013k} \end{tabular} \\ \hline
\textbf{Fusion Applied} & Simple Sum Fusion \\ \hline
\textbf{Modality Relationship Metrics} & \begin{tabular}[c]{@{}l@{}} Pearson Pairwise Correlation \\ Spearman Rank Correlation \end{tabular} \\ \hline
\end{tabular}%
}
\label{tab:experiments}
\end{table}

\subsection{Datasets}
This paper uses 3 multimodal datasets: NIST BSSR1 \cite{BSSR1nist}\footnote{NIST BSSR1 dataset is available at \url{https://www.nist.gov/itl/iad/image-group/nist-biometric-scores-set-bssr1}}, \biocop, and \trimodal \cite{mitlltrimodal}. In all these datasets, only the similarity scores are available, and only complete score vectors are analyzed. A brief description of these datasets is outlined below, and a summary is provided in Table \ref{dfs}.

\begin{figure}[htbp]
    \includegraphics[width=.25\textwidth]{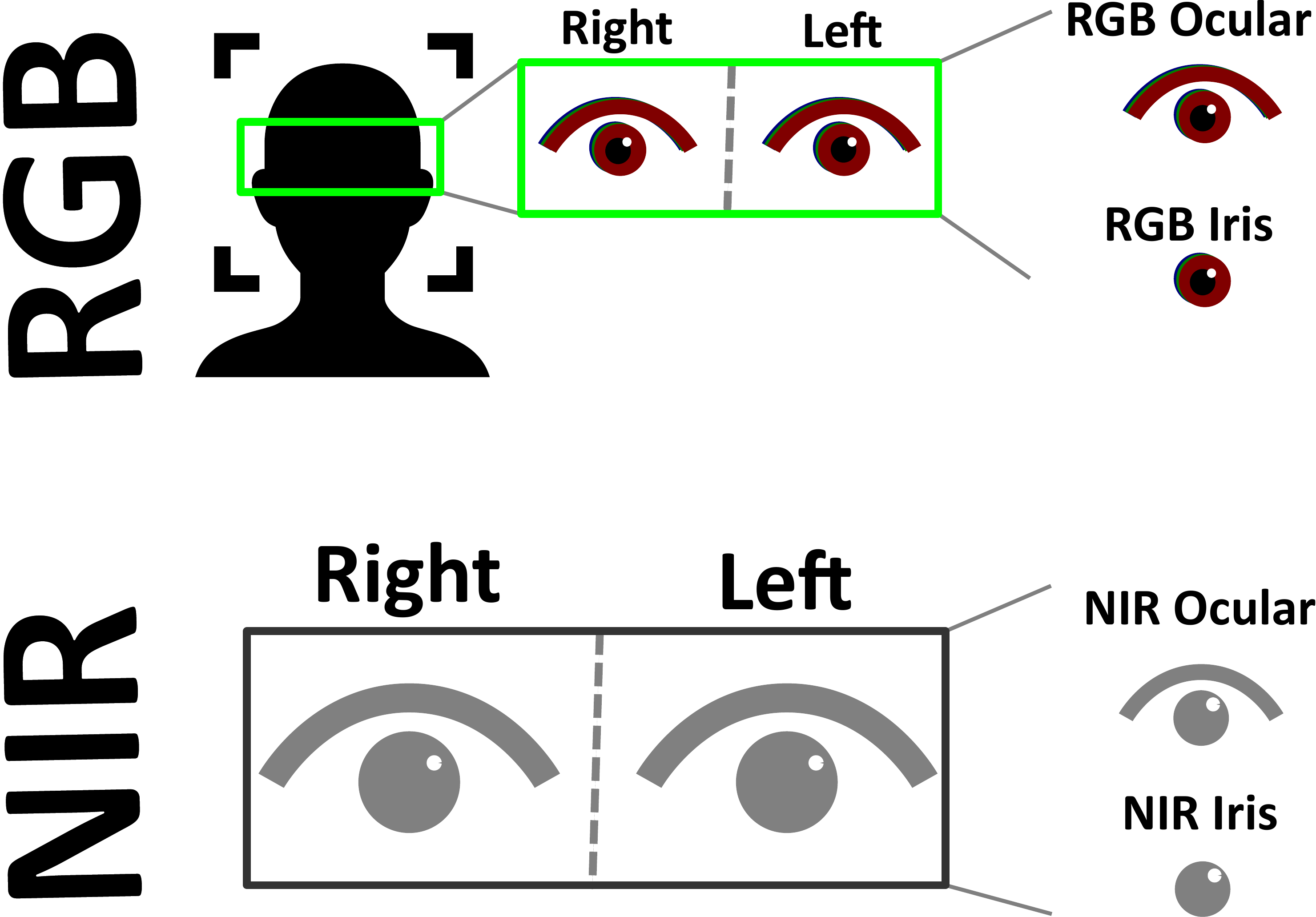}
    \centering
    \caption{Diagram of iris and ocular images cropped from ocular region in the \biocop dataset images (RGB facial images top, NIR ocular images bottom).}
    \label{fig:Ocular}
\end{figure}

The \biocop multimodal score dataset is produced by a multichannel CNN that performs cross-modal matching of face images with iris images. The CNN is trained on data collected from the BioCop 2008 ocular dataset, consisting of RGB face and NIR ocular images. From the RGB face images, the left and right eyes are cropped and aligned with the corresponding NIR images. Three copies of the NIR ocular image are stacked on top of the RGB image, creating a 6 channel patch which is then fed into a multichannel CNN to produce similarity scores. In addition to the ocular images for left and right eyes, the irises for each are cropped out to obtain a new type of score belonging to the iris (diagrammed in Figure \ref{fig:Ocular}). This process produces 4 types of scores that form the score vector between 2 subjects: left ocular, right ocular, left iris, and right iris. In this dataset, approximately 30\% of the score vectors contain naturally occurring missing score values.

The NIST BSSR1 multimodal dataset comprises similarity scores obtained from four modalities: left thumbprint, right thumbprint, face matcher C, and face matcher G. These scores are derived by comparing a subject's sample against the samples of identities in a gallery consisting of 517 subjects. Note, the provided face matcher scores come from matching algorithms NIST generically refers to as C and G.

The \trimodal{} multimodal dataset is created by collecting scores from high quality face and speech modalities present in the VoxCeleb-H dataset (referred to as the hard-set) \cite{nagrani2020voxceleb}. In addition to the face and speech modalities, a text modality is pulled from a subset of the PAN Celebrity Profiling Twitter dataset \cite{PAN}. Note, as opposed to the previous datasets where the amount of imposter score vectors vastly outnumber the genuine score vectors, the majority of the score vectors in the \trimodal{} dataset are genuine.

\begin{table}[htbp]
\centering
\caption{Summary of Multibiometric datasets analyzed.}
\label{dfs}
\resizebox{\columnwidth}{!}{%
\begin{tabular}{llll}
\toprule
{} & Modalities & Score Vectors &  Genuine \\
\midrule
\textbf{\biocop} & 4 & 139,230 & 435  (0.31\%) \\
\textbf{NIST BSSR1} & 4 & 133,903 & 517  (0.39\%) \\
\textbf{\trimodal} & 3 & 107,471 & 77,789  (72.38\%) \\
\bottomrule
\end{tabular}%
}
\end{table}

\subsection{Missing Score Simulations}

We simulate missing score data from 0\% missing up to 90\% missing. To simulate these missing scores, we first randomly select the score vectors to be corrupted. From each of these selected vectors, a random number between 1 and the number of modalities-1 of the vector's scores is dropped. This ensures that at least one score will be dropped and at most all-but-one score will be dropped. The pseudocode for this process is summarized in Algorithm \ref{ALG:rando}. This process is repeated 5 times on the train and test sets. For the \biocop dataset, which contains naturally occurring missing score values, we first drop the incomplete score vectors before simulation. Note, we provide the performance of the simulated missing values to the naturally occurring missing values in the Results section.

\begin{algorithm}
	\caption{Simulation of Missing Data}
    \label{ALG:rando}
	\begin{algorithmic}[1] 
		\For {$\text{proportion}=0, 10, 20, \ldots, 90$}
            \State $n = \text{Integer}\left(\frac{\text{proportion}}{100} \times \text{length(score data)}\right)$
			\State $corrupted = \text{random.sample}(n, \text{score data})$
            \For {$\text{vector} \in \text{corrupted}$}
                \State $\text{amount2drop} = \text{random}(1, \text{length(modalities)}-1)$
                \State $dropped = \text{random.sample}(\text{amount2drop}, \text{vector})$
                \State $\text{score data}[dropped] = \text{NaN}$
            \EndFor
		\EndFor
	\end{algorithmic} 
\end{algorithm}

We apply this process to multiple versions of each dataset: Genuine Missing, Imposter Missing, and Any Missing. For the \textit{Genuine Missing}, only the vectors belonging to the genuine label are altered to contain missing values. Likewise, the \textit{Imposter Missing} indicates only imposter vectors have been altered. \textit{Any Missing} contains randomly simulated missing values regardless of the label. 

Because the genuine and imposter classes are often unbalanced, we also consider a balanced training set for the above versions. For example, in the NIST BSSR1 dataset each subject id contains a score vector for every subject in the gallery. That is, for every 1 genuine score vector, there are 516 imposter score vectors, leading to the potential for over-fitting. We consider a balanced version of the training dataset, where we randomly down sample from the larger class to be the same size as the smaller class. Note that the results presented in Section \ref{sec:results} have been generated on the same test set. Tables \ref{tab:BioCop-sums}, \ref{tab:NIST BSSR1-sums}, and \ref{tab:Trimodal-sums} show the differences between the original dataset and the balanced dataset versions.

\begin{table}
\centering
\caption{\biocop Comparison of original and balanced dataset versions}
\label{tab:BioCop-sums}
\begin{tabular}{lll}
\toprule
{} & Original & Balanced \\
\midrule
 Score Vectors          &   139,230 &      870 \\
 Genuine Score Vectors  &      435 &      435 \\
 Imposter Score Vectors &   138,795 &      435 \\
\bottomrule
\end{tabular}
\end{table}

\begin{table}
\centering
\caption{NIST BSSR1 Comparison of original and balanced dataset versions}
\label{tab:NIST BSSR1-sums}
\begin{tabular}{lll}
\toprule
{} & Original & Balanced \\
\midrule
Score Vectors          &   133,903 &     1,034 \\
Genuine Score Vectors  &      517 &      517 \\
Imposter Score Vectors &   133,386 &      517 \\
\bottomrule
\end{tabular}
\end{table}

\begin{table}
\centering
\caption{\trimodal{} Comparison of original and balanced dataset versions}
\label{tab:Trimodal-sums}
\begin{tabular}{lll}
\toprule
{} & Original & Balanced \\
\midrule
Score Vectors          &   107,471 &    59,364 \\
Genuine Score Vectors  &    77,789 &    29,682 \\
Imposter Score Vectors &    29,682 &    29,682 \\
\bottomrule
\end{tabular}
\end{table}

\subsection{Imputation}
For our analysis, we consider the verification performance of simple sum fusion on the simulated missing dataset as the baseline performance. We then apply univariate mean and median imputations, as well as multivariate imputation through MICE with Bayesian, Decision Tree, and KNN regressor models.  \textcolor{black}{We additionally examine how reducing training data impacts imputation outcomes.} Note that none of the imputation techniques use vector labels (i.e., genuine or imposter), but rather are only calculated on scores within the modality (univariate) or scores across modalities (multivariate). These models are defined below. 






Univariate imputation approaches utilize only the scores of the missing score's modality from the training set. \textbf{Mean} imputation replaces missing scores within a modality with the mean score of the available scores for that modality in the training set. Similarly, \textbf{Median} imputation replaces missing scores within a modality with the median score of the available scores for that modality in the training set.

In addition to the above univariate approaches, we apply multivariate imputation methods using the MICE method (described in Section \ref{sec:Background}) with the following supervised models. It should be noted that emphasis was not placed on the parameter tuning in these models, and it is possible performance could be further improved with parameter optimization strategies. 

MICE with \textbf{Bayesian Ridge Regression} is a probabilistic model of regression $p(y \mid X, w, \alpha)=\mathcal{N}(y \mid X w, \alpha)$, where parameters are estimated by maximizing the log marginal likelihood.

MICE with \textbf{Decision Tree Regression} is a non-parametric model of regression. Decision Trees aim to learn a hierarchy of decision rules inferred from the training data's features. This approach can potentially be more resilient to violations in underlying model assumptions; however it can also be prone to over-fitting. 

MICE with \textbf{KNN Regression} imputation applies a K Nearest Neighbor (KNN) approach. Score vectors in the training data are sorted by distance, and the scores of the k-nearest neighbors (i.e. those with the smallest distances) are averaged. The experiments presented here set k to 5 neighbors and use the Euclidean distance to measure the distance of 2 points in the $m$ dimensional space.

\section{Results}
\label{sec:results}
\begin{figure}
     \centering
     \begin{subfigure}[b]{0.5\textwidth}
         \centering
         \includegraphics[width=\textwidth]{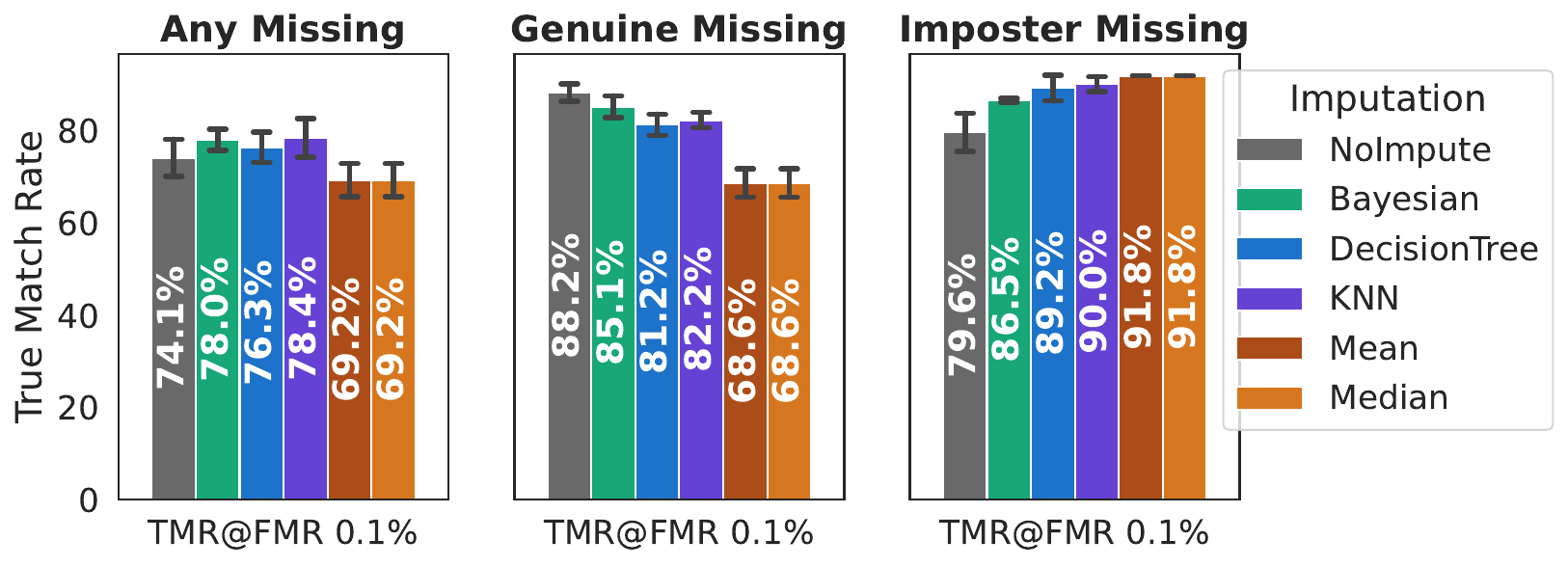}
         \caption{BioCop 2008 Score Set}
         \label{fig:y equals x}
     \end{subfigure}
     \hfill
     \begin{subfigure}[b]{0.5\textwidth}
         \centering
         \includegraphics[width=\textwidth]{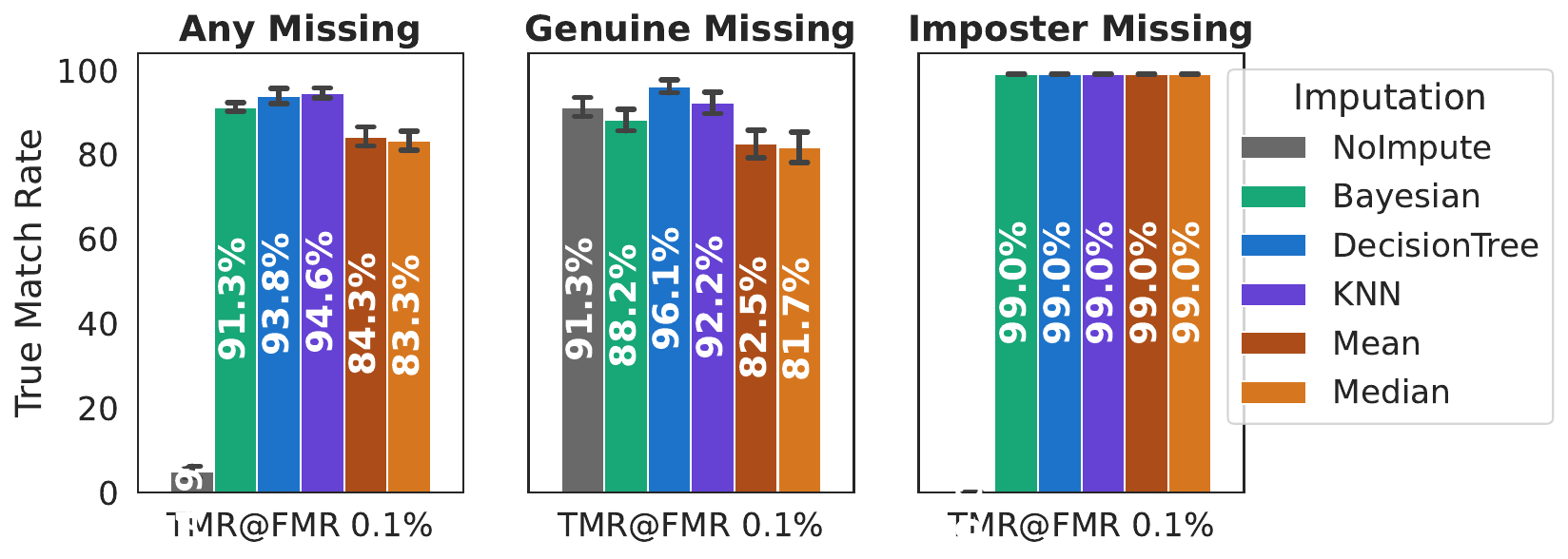}
         \caption{NIST BSSR1 Score Set}
         \label{fig:three sin x}
     \end{subfigure}
     \hfill
     \begin{subfigure}[b]{0.5\textwidth}
         \centering
         \includegraphics[width=\textwidth]{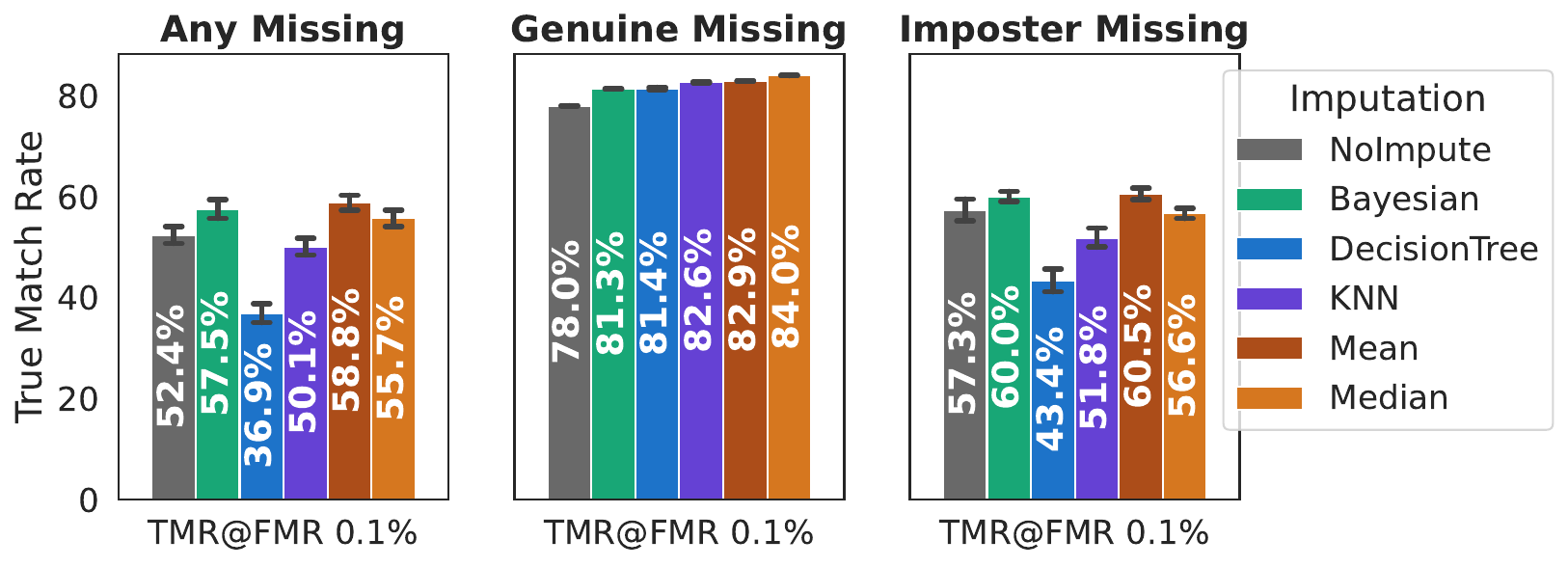}
         \caption{\trimodal{} Score Set}

     \end{subfigure}
        \caption{Estimated TMR at FMR=0.1\% for 50\% incomplete score vectors found in the unbalanced version of the training data.}
        \label{fig:verification}
\end{figure}

\begin{figure}
     \centering
     \begin{subfigure}[b]{0.48\textwidth}
         \centering
         \includegraphics[width=\textwidth]{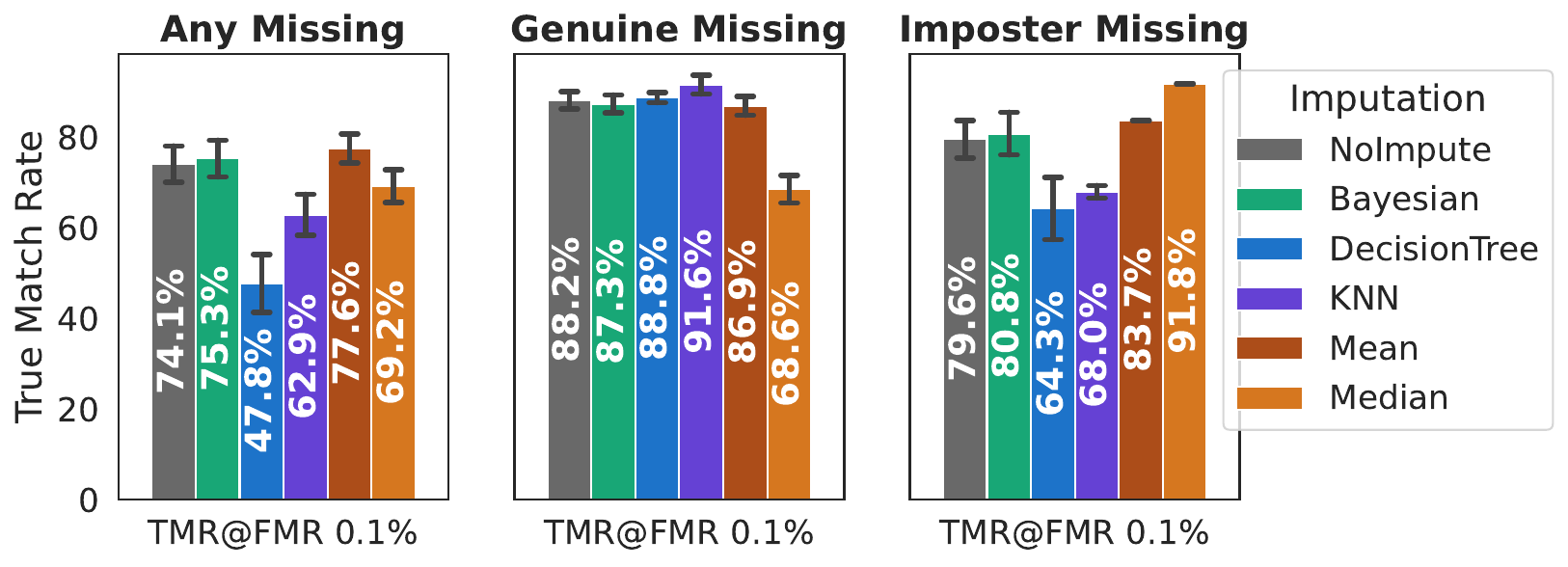}
         \caption{Balanced BioCop 2008 Score Set}
     \end{subfigure}
     \hfill
     \begin{subfigure}[b]{0.48\textwidth}
         \centering
         \includegraphics[width=\textwidth]{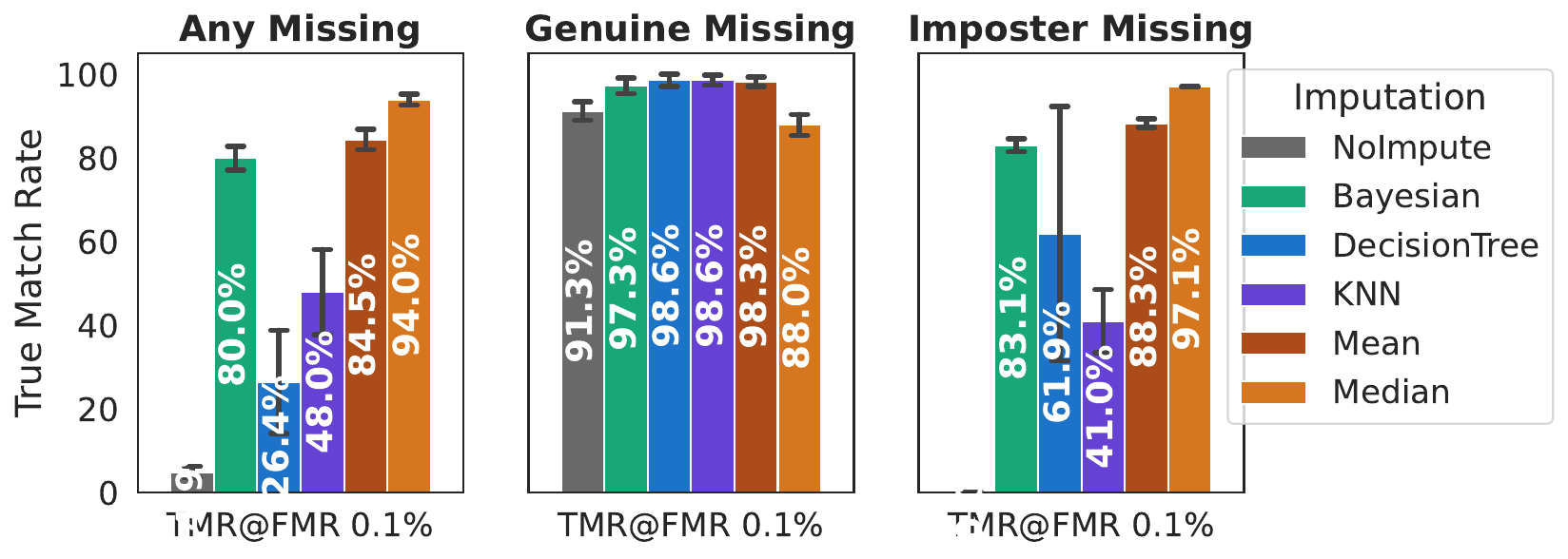}
         \caption{Balanced NIST BSSR1 Score Set}
     \end{subfigure}
     \hfill
     \begin{subfigure}[b]{0.48\textwidth}
         \centering
         \includegraphics[width=\textwidth]{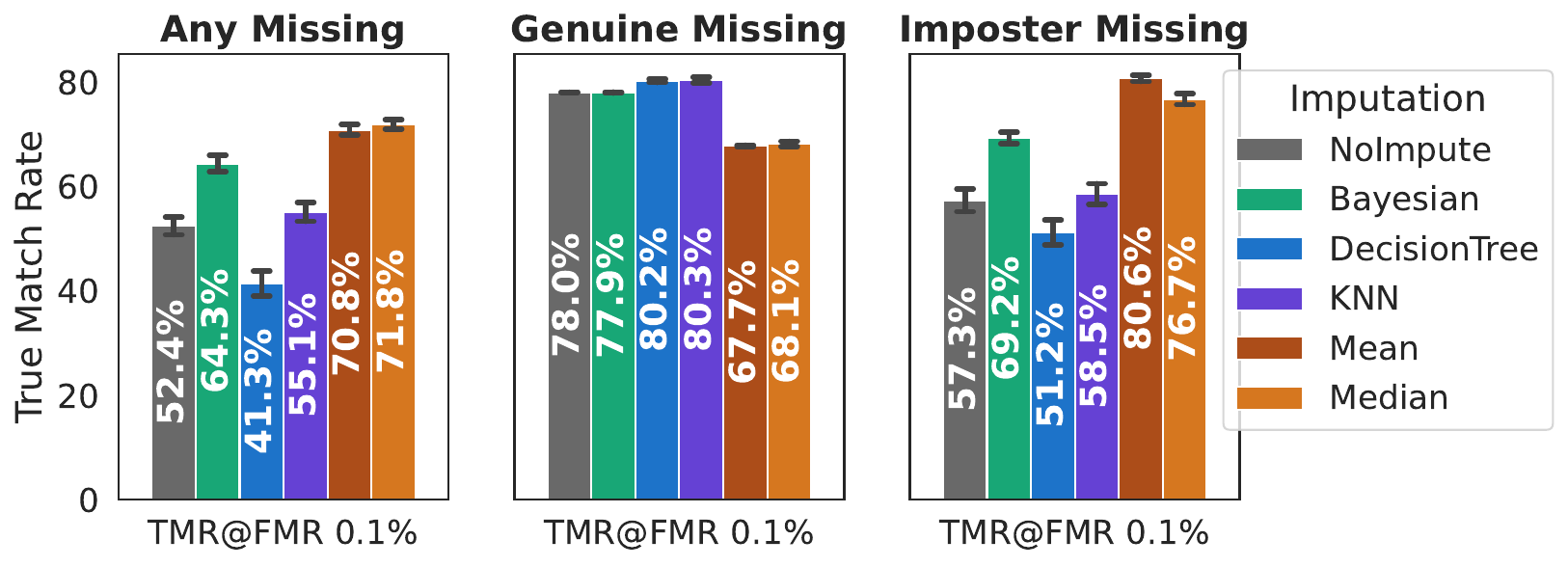}
         \caption{Balanced \trimodal{} Score Set}

     \end{subfigure}
        \caption{Estimated TMR at FMR=0.1\% for 50\% incomplete score vectors trained on the balanced training set.}
        \label{fig:verification-balanced}
\end{figure}

In this section we highlight a small subset of the results from the experimental approach described above. Here we highlight the verification performance at False Match Rate (FMR) = 0.1\% on the 50\% missing levels in Figures \ref{fig:verification} and \ref{fig:verification-balanced} \footnote{Complete results may be viewed at \url{https://melissadale.github.io/WIFS2023/}}.

In both the unbalanced and balanced versions, we see that fusion performance for the \textit{Any Missing} version the MICE with Bayesian Ridge Regression consistently improves performance over applying no imputation at all (even if it is not \textit{the best} imputation technique for each dataset). However we note that the verification performance of the test set varies between the approach trained on balanced data and the approach trained with the unbalanced data. In the unbalanced training approaches, the imputed genuine scores do not show such a clear gain for the \biocop and NIST BSSR1 datasets, where there are substantially fewer genuine scores to train imputation methods on. Conversely the \trimodal{} dataset contains a larger proportion of genuine scores in which to train imputation methods (Figure \ref{fig:verification}). This observation highlights the biases introduced by the overrepresented class in the training set. However, we also emphasize that despite the exclusion of a significant number of valid score vectors to achieve class balance, the overall verification performances are not severely compromised. \textcolor{black}{While more training samples are generally preferred in machine learning, our observations remain consistent regardless of training data size.}

Our analysis reveals that multivariate methods surpass univariate ones in imputing missing genuine scores. Conversely, mean or median imputation proves more effective for imposter scores. We note that the genuine class displays a more pronounced average pairwise correlation than the imposter class, potentially illuminating  MICE's superior performance with the genuine class. This underscores the role the majority class plays in shaping both the overall average pairwise correlation and importantly the resulting imputation outcomes.



\subsection{Missing Simulation versus Naturally Occurring Missing}
In our experiments, we simulated random missing scores from complete versions of each dataset to meet the Missing at Random (MAR) requirements. However, it is important to highlight that the verification performance of naturally occurring missing scores within the \biocop dataset is comparable to the performance reported in the simulated missing data, as presented in Table \ref{tab:irl}. This demonstrates the robustness and relevance of our findings, indicating that our results hold promise for real-world scenarios.  

\begin{table}[htbp]
\caption{Comparison of performance observed in naturally occurring missing value and simulated missing values observed in the \biocop dataset.}
\label{tab:irl}
\begin{tabular}{|l|c|c|}
\hline
                           & \multicolumn{1}{l|}{\textbf{Amount Missing}} & \multicolumn{1}{l|}{\textbf{TMR@FMR=0.1\%}} \\ \hline
\textbf{Naturally Missing} & 30.39\%                                      & 81.68                                       \\ \hline
\textbf{Simulated Missing} & 30\%                                         & 80.20 (+/- 5.0)                             \\ \hline
\end{tabular}
\end{table}

\section{Conclusion and Discussion}
In this study, we investigated the influence of imputation techniques on verification performance in multibiometric score datasets. Our findings have important implications for improving the accuracy and reliability of multibiometric systems with missing scores. We summarize key observations and the resulting recommendations below. 

Firstly, our results consistently demonstrate the benefits of imputation over not imputing missing scores, regardless of the type of scores being imputed. By incorporating imputation into the data preprocessing stage, multibiometric systems can benefit from more complete score data, thereby improving overall system performance. \textbf{Recommendation 1: } \textit{Enhance multibiometric system design by integrating imputation techniques.} Invest time in finding the most appropriate approach for your data.

Secondly, we observed that imputation methods tend to favor the overrepresented class, introducing biases in the imputed scores. To mitigate this issue, we emphasize the importance of balancing the classes within the training dataset. Despite the potential need to drop a substantial number of data points from the overrepresented class, our analysis demonstrates that this step does not severely harm the overall verification performance. Balancing the training data helps alleviate biases and ensures fair representation of both genuine and imposter scores, leading to more reliable and unbiased performance evaluation. \textbf{Recommendation 2: } \textit{Balance the imposter and genuine score vectors in the training set.} Although balancing the training data may involve disregarding a significant number of score vectors from the overrepresented class, it is important to note that the performance of the balanced versions in the test set is not significantly compromised for the datasets analyzed. 

Furthermore, our study highlights the effectiveness of different imputation approaches based on the class of the missing scores (genuine or impostor). An analysis of the mean pairwise correlation between biometric modalities can also provide insights into the choice of imputation approach. \textbf{Recommendation 3: } \textit{When designing an imputation approach, consider the nature of missing scores and any inherent correlations between between scores of modalities.} Understanding which types of scores are more prone to being missing, along with inter-modality relationships, can inform the development of targeted and effective imputation strategies. This observation suggests future research efforts are required to develop a novel approach to imputation in multibiometrics. By creating methods to manage missing scores without prior label knowledge, we can boost recognition accuracy and strengthen practical biometric applications. 


In conclusion, our study provides insights into the role of imputation techniques in multibiometric score datasets. By leveraging imputation, multibiometric systems can enhance their recognition accuracy and reliability. Balancing the training data and employing appropriate imputation methods based on score type are essential considerations for achieving optimal performance. Our findings contribute to the understanding of imputation in the context of multibiometric systems and pave the way for future research in this area.

\section{Future Work}
While this study has shed light on the role of imputation techniques in multibiometric score datasets, there are several avenues for future research in this area. One important direction for future work is the development of innovative imputation methods that can effectively handle missing scores without relying on prior knowledge of the label but in a dynamic fashion. This includes investigating hybrid approaches that combine multiple imputation techniques or incorporate other data preprocessing methods. Hybrid methods have the potential to leverage the strengths of different imputation techniques and improve overall system performance. Additionally, although balancing the training data was found to mitigate biases introduced by imputation methods, further investigation is needed to explore more advanced techniques for addressing class imbalance. Future research can explore methods such as oversampling, undersampling, or generating synthetic data to ensure fair representation of both genuine and imposter scores in the training dataset.

\bibliographystyle{./IEEEtran}
\bibliography{./IEEEtran}

\end{document}